\documentclass[conference]{IEEEtran}
\IEEEoverridecommandlockouts
\usepackage{cite}
\usepackage{tikz}
\usepackage{amsmath,amssymb,amsfonts}
\usepackage{algorithmic}
\usepackage{graphicx}
\usepackage{textcomp}
\usepackage{xcolor}
\usepackage{booktabs} 

\newcommand{\ignore}[1]{}
\newcommand{\Name}{LLM-ProS}
\def\BibTeX{{\rm B\kern-.05em{\sc i\kern-.025em b}\kern-.08em
    T\kern-.1667em\lower.7ex\hbox{E}\kern-.125emX}}

\begin{document}

\title{LLM-ProS: Analyzing Large Language Models' Performance in Competitive Problem Solving}
\author{\IEEEauthorblockN{Md Sifat Hossain, Anika Tabassum, Md. Fahim Arefin}
\IEEEauthorblockA{
Department of Computer Science and Engineering,\\
University of Dhaka, Dhaka, Bangladesh \\
Email: $\lbrace$\textit{mdsifat-2019217800@cs.du.ac.bd,
}\\
\textit{anika-2019417844@cs.du.ac.bd,}\\
\textit{fahim@cse.du.ac.bd}$\rbrace$}\\
\and
\IEEEauthorblockN{Tarannum Shaila Zaman}
\IEEEauthorblockA{
Department of Information Systems, \\
University of Maryland, Baltimore County, Maryland, USA\\
Email: $\lbrace$\textit{zamant@umbc.edu}$\rbrace$}
}

\newcommand\copyrighttext{%
  \footnotesize \textcopyright 2025 IEEE. Personal use of this material is permitted.
  Permission from IEEE must be obtained for all other uses, in any current or future
  media, including reprinting/republishing this material for advertising or promotional
  purposes, creating new collective works, for resale or redistribution to servers or
  lists, or reuse of any copyrighted component of this work in other works.}
  
\newcommand\copyrightnotice{%
\begin{tikzpicture}[remember picture,overlay]
\node[anchor=south,yshift=10pt] at (current page.south) {\fbox{\parbox{\dimexpr\textwidth-\fboxsep-\fboxrule\relax}{\copyrighttext}}};
\end{tikzpicture}%
}

\maketitle
\copyrightnotice

\begin{abstract}

The rapid advancement of large language models has opened new avenues for automating complex problem-solving tasks such as algorithmic coding and competitive programming. This paper introduces a novel evaluation technique, LLM-ProS, to assess the performance of state-of-the-art LLMs on International Collegiate Programming Contest (ICPC) problems. Using a curated dataset of 166 World Finals problems from 2011 to 2024, we benchmark the models' reasoning, accuracy, and efficiency. We evaluate the five models-GPT-4o, Mistral Large, Llama-3.1-405B, and the o1 family, consisting of o1-mini and o1-preview, across critical metrics like correctness, resource utilization, and response calibration. Our results reveal significant differences in the models' abilities to generalize, adapt, and solve novel problems. We also investigated the impact of training methodologies, dataset contamination, and chain-of-thought reasoning on model performance. The findings provide new insights into optimizing LLMs for algorithmic tasks, highlighting both strengths and limitations of current models.

\end{abstract}


\section{Introduction}
Large language models (LLMs)\cite{b14} have revolutionized natural language processing (NLP)\cite{b15} and their applications in various technical domains, including algorithmic problem-solving and code generation. Trained on extensive and diverse datasets, these models exhibit remarkable capabilities in understanding and generating code\cite{b19}. However, their performance on complex, real-world coding challenges remains an area of active investigation. Competitive programming problems, such as those from the International Collegiate Programming Contest (ICPC)\cite{b8}, offer a unique opportunity to rigorously evaluate LLMs due to their intricate constraints, computational demands, and emphasis on algorithmic efficiency.

ICPC problems are particularly well-suited for evaluating LLM performance\cite{Austin2021Program} for several reasons. First, these problems require a combination of logical reasoning, algorithmic thinking, and precise implementation, reflecting real-world software engineering challenges. Second, their diverse categories, including graph theory, dynamic programming, and computational geometry, offer a broad spectrum of difficulty levels, testing an LLM's\cite{Carlini2022Quantifying} ability to generalize across various problem types. Third, the focus on correctness, efficiency, and edge-case handling aligns with critical evaluation metrics such as runtime performance and memory usage \cite{b1,b2,b10}.

Existing works use LLMs at different stages of code generation and testing. For example, Fakhoury et al. \cite{b3} improve solutions through test feedback and refine them step-by-step. Liu et al. \cite{b1} identify problems like data contamination and over-reliance on memorized patterns, which reduce the reliability of results. A recent study \cite{b13} explores how adding syntax and grammar rules during training can enhance the accuracy and robustness of LLMs. Similarly, Coignion et al. \cite{b27} analyze the efficiency of code generated by LLMs on Leetcode, revealing performance trends and benchmarks. However, these studies primarily focus on specific aspects of code generation, such as correctness or efficiency, without comprehensively evaluating LLM performance in solving complex, real-world competitive programming problems. To address this gap, we propose an approach, \Name{}, to assess the capabilities of advanced LLMs, including OpenAI’s o1 family\cite{b18}, GPT-4o\cite{b16}, Mistral Large\cite{b7}, and Llama-3.1-405B\cite{b17}, in tackling ICPC problems. 

To develop \Name{}, we collect a curated dataset of 166 ICPC World Finals problems from the years 2011 to 2024, providing a comprehensive benchmark for assessing the reasoning, accuracy, and efficiency of LLMs. The study assessed five state-of-the-art models—GPT-4o, Mistral Large, Llama-3.1-405B, o1-preview and o1-mini-selected for their diverse architectures and training methodologies. Comprehensive data preprocessing involved extracting problem components, standardizing prompts, cleaning text, customizing templates for each model, and validating the preprocessed data. Solutions generated by the LLMs were submitted to Codeforces Gym ICPC contests, receiving automated feedback on correctness and efficiency. All experiments were conducted in a controlled environment, with reproducible submissions and publicly available scripts to ensure consistency and transparency. In summary we make the following contributions:
\begin{itemize}
    \item We propose a performance analyzer, \Name{}, to assess the performance of five different LLMs on new ICPC problems and examine accuracy trends over time.
    \item We present an experimental evaluation that identifies factors contributing to variability in LLM performance and highlights the distribution of verdicts across various LLMs for the given problem set.
\end{itemize}

Our findings reveal that o1 models, due to their enhanced Chain of Thought reasoning and calibration, significantly outperform others in terms of accuracy, robustness, and efficiency. Additionally, we identify critical limitations, such as overestimation of confidence by some models and their struggle with complex, high-difficulty problems. Recent studies, such as \emph{The Llama 3 Herd of Models} \cite{b11}, further highlight the challenges in ensuring consistent performance across diverse technical domains, emphasizing the need for specialized benchmarks. By addressing these issues and leveraging insights from ICPC problem evaluations, this paper lays the foundation for improving LLM design and training methodologies to meet the challenges of technical problem-solving.

\section{Background and Motivation}

\subsection{Motivation}
Despite their remarkable capabilities, the effectiveness of LLMs in tackling complex, real-world coding challenges remains inadequately explored. Traditional benchmarks often fall short in assessing the nuanced reasoning, adaptability, and efficiency required for high-stakes programming tasks.

The International Collegiate Programming Contest (ICPC), the oldest, largest, and most prestigious programming contest in the world, presents a unique and rigorous environment for evaluating LLM performance\cite{Chen2021Evaluating}. ICPC-style problems are characterized by their intricate constraints, diverse problem categories, and emphasis on algorithmic efficiency and edge-case handling. These attributes make them ideal for benchmarking the reasoning capabilities, accuracy, and resource utilization of LLMs in a controlled yet challenging setting \cite{b1,b2}.

Moreover, the increasing reliance on LLMs for automated code generation in software engineering necessitates a deeper understanding of their strengths and limitations. Evaluating LLMs against competition-level programming problems not only highlights their problem-solving prowess but also identifies critical areas for improvement, such as generalization to unseen problems and efficient resource usage \cite{b3,b4}. This study aims to bridge the existing research gaps by providing a comprehensive assessment of state-of-the-art LLMs using a curated set of ICPC World Finals problems.

\subsection{Background}
To thoroughly evaluate the performance of LLMs on ICPC-style problems, we selected a diverse range of models representing various architectures, training methodologies, and optimization strategies. The models under consideration include GPT-4o, Mistral Large, Llama-3.1-405B, o1-mini, and o1-preview. Each model was chosen for its distinct characteristics and potential applicability to structured problem-solving tasks.

\subsubsection{GPT-4o}
GPT-4o \cite{b27,b11,b26} is a general-purpose language model renowned for its high accuracy in code generation and reasoning tasks. Leveraging its extensive training on diverse datasets, GPT-4o\cite{Extending} demonstrates strong capabilities in understanding complex problem statements and generating syntactically correct code. However, its performance may be less optimized for structured problem-solving compared to models specifically fine-tuned for reasoning and iterative refinement.

\subsubsection{Mistral Large}
Mistral Large specializes in handling domain-specific tasks with a focus on efficient resource utilization. Designed to balance performance with computational efficiency, Mistral Large offers insights into the trade-offs between model complexity and practical deployment in resource-constrained environments. Its performance on ICPC problems provides valuable data on how domain specialization impacts problem-solving efficacy \cite{b13}.

\subsubsection{Llama-3.1-405B}
Llama-3.1-405B \cite{b27,b12} is a computationally efficient model tailored for general-purpose use. Serving as a benchmark for lightweight LLMs in competitive programming, Llama-3.1-405B facilitates comparisons regarding scalability and efficiency without significant compromises in accuracy. Its performance metrics shed light on the feasibility of deploying smaller models in high-stakes programming scenarios. It is also particularly interesting as the best-performing open source model.

\subsubsection{OpenAI o1 Family}
The o1-mini and o1-preview models \cite{b4,b10} from the OpenAI o1 family are specifically fine-tuned for chain-of-thought (CoT) reasoning and iterative refinement processes. These models are engineered to excel in structured, multi-step problem-solving tasks, making them highly suitable for ICPC-style evaluations. Their design emphasizes enhanced reasoning pathways and reduced hallucination rates, enabling more consistent and accurate code generation under complex constraints.

By systematically applying each model to this diverse set of problems, the study evaluates not only the accuracy and robustness of the solutions but also the computational efficiency and adaptability of the models in handling unseen and complex programming tasks. This rigorous evaluation framework provides a holistic view of the current capabilities of LLMs in competitive programming contexts, informing future developments in model design and training methodologies.

\begin{figure*}[htbp]
\centerline{\includegraphics[scale=0.55]{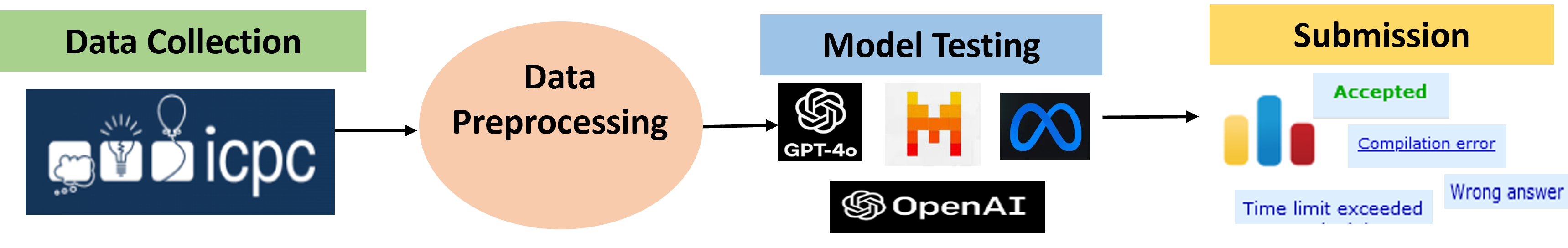}}
\caption{The Overview of \Name{}}
\label{fig:overview}
\end{figure*}
\section{Methodology}
Figure \ref{fig:overview} illustrates the overview of our proposed technique, \Name{}. \Name{} consists of four major steps: data collection, data preprocessing, model testing, and solution generation and submission. The details of each step are discussed in the following subsections.

\subsection{Data Collection}
We scrape a total of 166 competitive programming problems from the ICPC official website \cite{b8}, specifically from the World Finals editions spanning the years 2011 to 2024 World Finals. We select these years to avoid potential overlap with the training data of major LLMs, ensuring a more accurate evaluation of their performance. Table \ref{tab:problemst} shows a sample problem structure that we have collected. Each problem includes comprehensive details such as problem statements, input formats, output formats, sample inputs and outputs, notes, and specified time and memory limits. This dataset provides a diverse and challenging set of tasks that effectively test the reasoning and computational abilities of LLMs.
\begin{table}[htbp]
\centering
\caption{Sample Problem Data}
\label{tab:problemst}
\begin{tabular}{|l|p{5cm}|}
\hline
\textbf{Problem Statement} & Allied Chute Manufacturers is a company that builds trash chutes. A trash chute ... \\
\hline
\textbf{Input-Output} & The input contains several test cases. Each test case starts with an integer $n$, representing the number of points in the polygon that models the trash item ($3 \leq n \leq 100$)... \\
\hline
\textbf{Sample testcase} & 
\textbf{Sample Input:} \newline
\texttt{3} \newline
\texttt{0 0} \newline
\texttt{...} \newline
\textbf{Sample Output:} \newline
\texttt{Case 1: 2.40} \newline
\texttt{Case 2: 14.15} \\
\hline
\end{tabular}
\end{table}

\subsection{Data Preprocessing}
Before generating solutions, we systematically preprocess each of the 166 ICPC World Finals problems to ensure compatibility and consistency across different LLMs. The preprocessing steps include:\\
\textbf{Extraction of Problem Components}: We break down each problem into its fundamental components, including the problem statement, input format, output format, sample inputs and outputs, notes, and specified time and memory limits. This structured extraction helps us create comprehensive prompts for the models.\\
\textbf{Standardized Prompt Formatting}: We format the extracted components into a consistent template to maintain uniformity across all prompts. The template includes sections such as ``Problem Statement", ``Input", ``Output", and ``Sample Test Cases", ensuring that each prompt provides all the necessary information for the model to generate an accurate solution.\\
\textbf{Text Cleaning and Normalization}: We clean all textual data to remove extraneous characters, ensure proper encoding, and maintain clarity. This step includes correcting formatting inconsistencies, standardizing terminology, and ensuring that mathematical notations and symbols are appropriately represented.\\
\textbf{Template Customization for Each Model}: Recognizing the unique characteristics and requirements of each LLM, we slightly adjust the templates to optimize prompt comprehension. For example, we provide models fine-tuned for chain-of-thought reasoning, such as o1-mini and o1-preview, with additional guiding instructions to facilitate step-by-step solution generation.\\
\textbf{Validation of Preprocessed Data}: We manually review the preprocessed prompts to ensure that all relevant information is accurately captured and that the prompts adhere to the standardized format. This validation step helps us identify and correct any discrepancies or omissions.

\subsection{Model Testing}
We evaluate five state-of-the-art LLMs: GPT-4o, Mistral Large, Llama-3.1-405B, and the OpenAI o1 Family, which consists o1-mini and o1-preview. Each model represents a different architecture, training methodology, and reasoning capability. We apply all five models to our preprocessed data. We select these models for their diverse training approaches, which represent a mix of general-purpose and task-optimized LLMs. We test them in a pass@1 setting to assess their ability to generalize to ICPC-style problems \cite{b1,b2}.

\ignore{
\subsubsection{GPT-4o}
GPT-4o \cite{b27} is a general-purpose model known for high accuracy in code generation and reasoning tasks but is less optimized for structured problem-solving compared to the o1 models.

\subsubsection{Mistral Large}
Mistral Large \cite{b11} specializes in handling domain-specific tasks with efficient resource usage. Its performance on ICPC problems provides insights into model efficiency versus generality.

\subsubsection{Llama-3.1-405B}
Llama-3.1-405B \cite{b10} is a smaller, computationally efficient model designed for general-purpose use, serving as a benchmark for lightweight LLMs in competitive programming.

\subsubsection{OpenAI o1 Family}
The o1-mini and o1-preview models \cite{b4} are fine-tuned for chain-of-thought (CoT) reasoning and iterative refinement, making them well-suited for solving structured, multi-step problems.
}

\subsection{Solution Generation and Submission}
We generate solutions to the scraped ICPC problems using the selected LLMs through their respective API keys. The generated solutions, which include code implementations aligned with the provided problem statements and constraints, are then submitted to the Codeforces Gym ICPC section contests \cite{b8}. Codeforces Gym provides automated feedback on correctness and efficiency, with verdicts such as ``Accepted" (AC), ``Wrong Answer" (WA), ``Time Limit Exceeded" (TLE), ``Runtime Error" (RE), and ``Compile Error" (CE). Submissions to Codeforces Gym are logged for reproducibility.\cite{b24}.
\section{Experimental Evaluation}
We use Selenium\cite{b21} for the automated downloading of PDFs from the website. For data pre-processing, we utilized PyPDF2\cite{b22} to extract text and re (regular expressions)\cite{b23} for cleaning and formatting the extracted problem statements. We evaluate five large language models (LLMs): GPT-4o \cite{b27}, Mistral Large \cite{b11}, Llama-3.1-405B \cite{b10}, and both o1-mini and o1-preview \cite{b4}. Solutions\cite{b25} generated using their respective APIs \cite{b9} are submitted to Codeforces Gym ICPC contests \cite{b8}. To evaluate \Name{}, we consider four research questions:\\
\textbf{RQ1:} How do LLMs perform on new ICPC problems
compared to those potentially seen during training?\\
\textbf{RQ2:} What are the trends in accuracy across different
categories?\\
\textbf{RQ3:} What factors contribute to variability in LLM performance?\\
\textbf{RQ4:} How are verdicts distributed across various LLMs
for the given problem set?

\subsection{Evaluation Metrics}
The performance of each model was assessed using the following metrics:\\
\textbf{Accuracy (pass@1)}: The percentage of problems solved correctly on the first attempt served as the primary metric of success \cite{b1,b2}.\\
\textbf{Error Analysis}: Errors were categorized based on verdict types to identify systematic failures in each model's reasoning or implementation \cite{b27}.\\
\textbf{Resource Utilization}: Average runtime and memory usage were measured by Codeforces to assess computational efficiency \cite{b4}.\\
\textbf{Temporal Comparison}: Models were tested on two subsets of data:
    \begin{itemize}
        \item \textbf{Past ICPC Problems}: Problems likely included in the models’ training data.
        \item \textbf{Unseen 2024 Problems}: Problems released after the models' training cutoff dates, ensuring evaluation on novel data \cite{b2,b27}.
    \end{itemize}

\subsection{Results \& Analysis}

\begin{figure}[htbp]
    \centering
    \includegraphics[scale= 0.38]{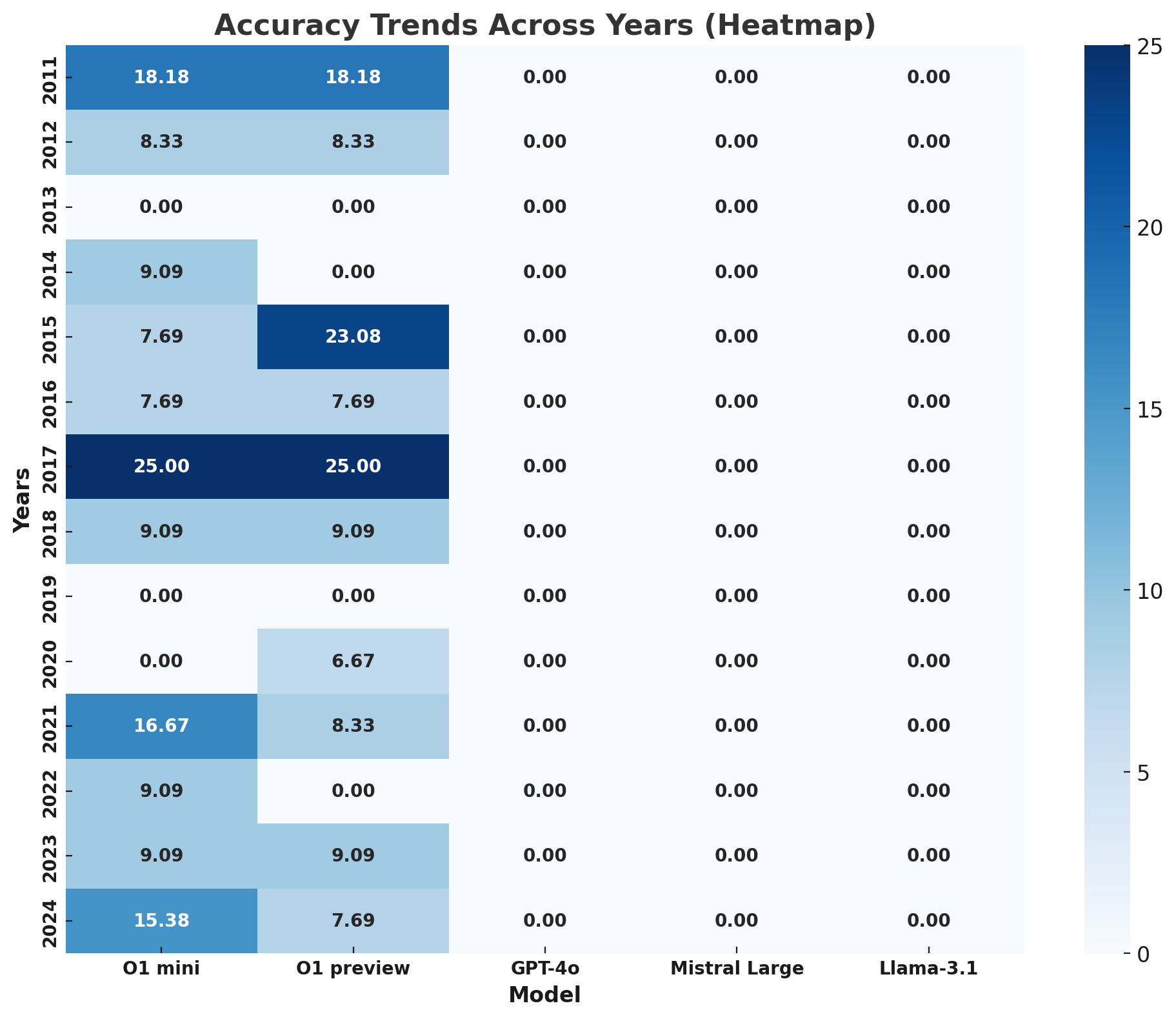}
    \caption{Accuracy Trends Across Years (Heatmap)}
    \label{fig:accuracy_heatmap}
\end{figure}

Evaluating large language models (LLMs) on ICPC problems reveals notable differences in accuracy, verdict distribution, and resource utilization across models. These findings offer detailed insights into our research questions, highlighting each model's strengths and limitations.\\
\textbf{RQ1: LLMs Performance on Unseen ICPC Problems:}
Figure~\ref{fig:accuracy_heatmap} presents a heatmap showing the accuracy trends of various LLMs across ICPC World Finals problems from 2011 to 2023 and the unseen 2024 dataset. The results highlight the stark contrast in performance between models optimized for reasoning and general-purpose models when handling challenging unseen problems.

The heatmap reveals that the o1-mini and o1-preview models demonstrate varied accuracy across the years, with both achieving their peak accuracy of 25.0\% in 2017. On the unseen 2024 dataset, o1-mini and o1-preview achieved accuracies of 15.4\% and 7.7\%, respectively, indicating their ability to generalize to new problem sets effectively.

In contrast, GPT-4o, Mistral Large, and Llama-3.1 consistently show 0\% accuracy across all years, including the unseen 2024 problems. These results emphasize their limitations in solving complex ICPC World Finals problems. This trend aligns with the hypothesis that general-purpose models lack the structured reasoning pathways necessary to excel in competition-level problem-solving without specific optimization.

Overall, the results confirm that models like o1-mini and o1-preview exhibit superior robustness and adaptability. Their consistent performance across years highlights the effectiveness of their fine-tuning and chain-of-thought reasoning capabilities. Conversely, the failure of other models to achieve any meaningful accuracy underscores the challenges faced by general-purpose LLMs in high-stakes competitive programming tasks.
\\
\textbf{RQ2: Accuracy Trends For Different Categories:}
The accuracy trends across problem categories provide insights into the strengths and limitations of the evaluated models. Categories like Implementation (5 problems), Graph Theory (3 problems) and Math (4 problems)\cite{Cobbe2021Training} saw higher overall success rates, particularly for the o1-mini and o1-preview models. These categories test the models' ability to handle structured and logical problem-solving tasks.

In contrast, categories such as Geometry and Greedy, each with only 1 problem, presented challenges across all models, indicating potential difficulties in handling specialized or less frequent problem types. This disparity highlights the need for targeted optimization to improve performance in these areas.

The following chart (Figure~\ref{fig:category_trends}) illustrates the number of solved problems across different categories:

\begin{figure}[htbp]
    \centering
    \includegraphics[width=0.45\textwidth]{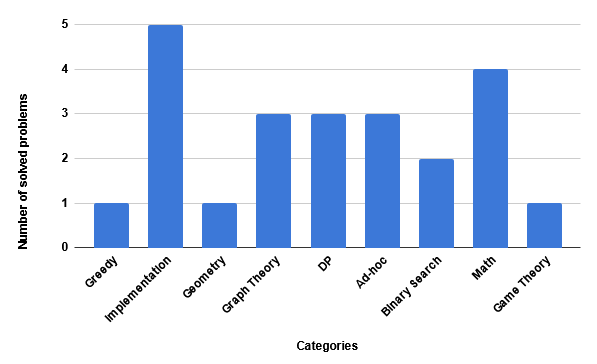}
    \caption{Number of solved problems across categories.}
    \label{fig:category_trends}
\end{figure}

The findings emphasize that problem category distribution plays a significant role in shaping model performance. The o1 models consistently demonstrated better adaptability across diverse categories compared to GPT-4o, Mistral Large, and Llama-3.1-405B, which struggled to generalize effectively. These trends align with the hypothesis that specialized training, such as chain-of-thought reasoning, enhances model capabilities in handling category-specific complexities.
\\
\textbf{RQ3: Variabilities in Model Performance:} Three key factors—dataset contamination, training methodologies, and reasoning strategies—contribute to variations in model performance.
\subsubsection{Dataset Contamination}
One major variability in LLM performance stems from dataset contamination. As shown in Figure~\ref{fig:accuracy_heatmap}, the accuracy of o1-mini and o1-preview on non-contaminated data (2024) is significantly lower than their highest accuracy on potentially contaminated data from earlier years. For example, o1-preview achieves 25.0\% accuracy in 2017, whereas its accuracy on the unseen 2024 problems drops to 7.7\%. Similarly, o1-mini's accuracy drops from 25.0\% in 2017 to 15.4\% in 2024. This disparity suggests that data contamination likely contributes to inflated performance metrics, as models may rely on memorized patterns rather than genuine reasoning for problems included in their training data. These results highlight the need for contamination-free datasets to accurately evaluate reasoning and generalization capabilities \cite{b1, b27}.

\subsubsection{Training Methodologies}
The training methodologies adopted by different models significantly influence their performance variability. As highlighted in Figure~\ref{fig:accuracy_heatmap} and Figure~\ref{fig:verdict_distribution}, o1-mini and o1-preview consistently outperform general-purpose models such as GPT-4o and Llama-3.1-405B. This success is attributed to their specialized fine-tuning, which emphasizes chain-of-thought (CoT) reasoning and iterative refinement.

Unlike general-purpose models, the o1-mini and o1-preview models are trained to simulate step-by-step problem-solving, enabling them to break down complex ICPC problems into manageable sub-tasks\cite{Patel2021Are}. For example, the verdict distribution analysis in Table~\ref{tab:verdict_distribution_counts} shows that o1-mini achieves 16 Accepted (AC) solutions and 10 Compile Errors (CE), compared to GPT-4o, which recorded 0 ACs and 41 CEs. These results highlight how tailored training methodologies enhance the reasoning pathways of o1 models, directly translating to higher success rates on unseen, complex tasks \cite{b4, b13}.

\subsubsection{Reasoning Strategies}
\begin{figure}[htbp]
    \centering
    \includegraphics[scale=0.35]{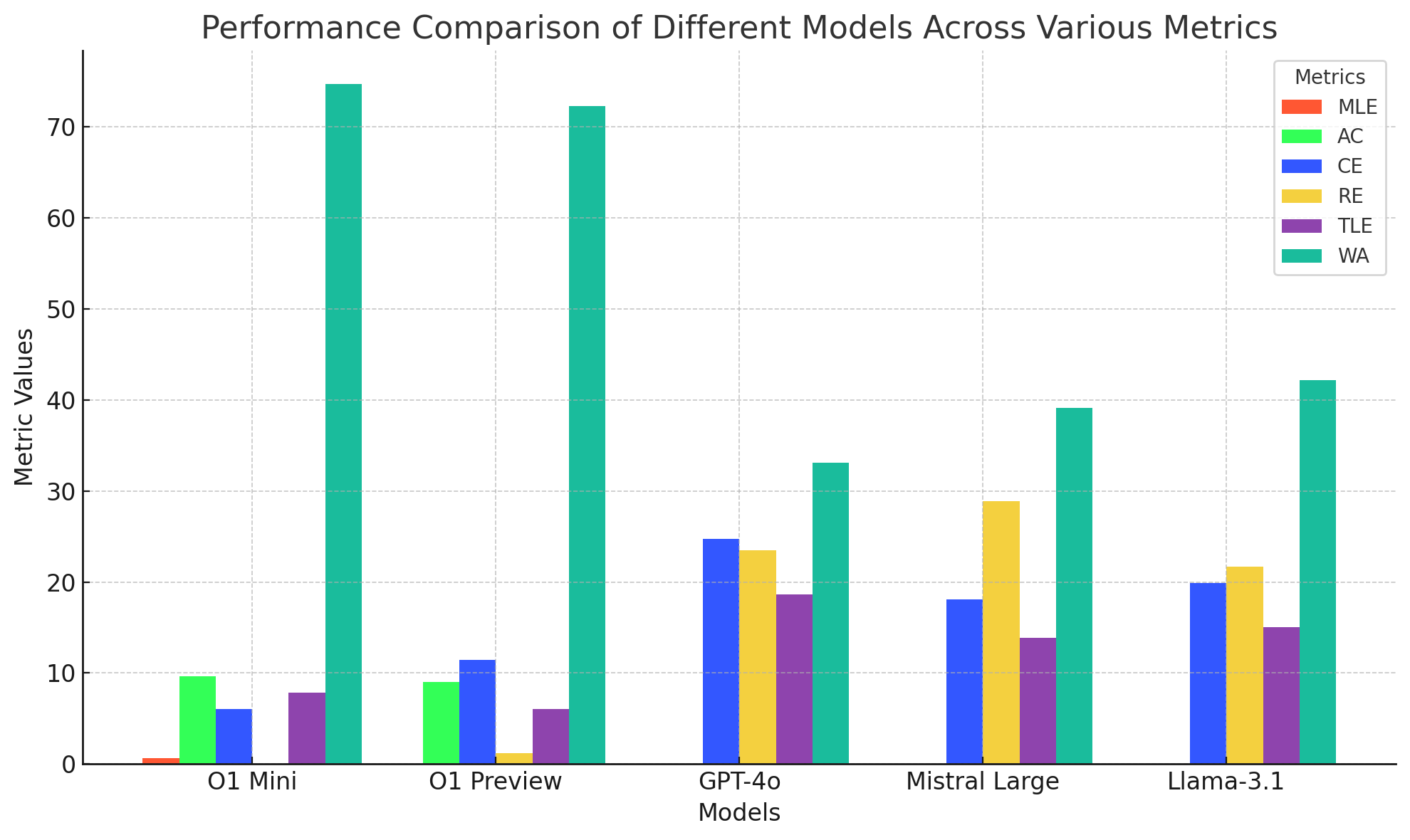}
    \caption{Verdict Distribution Across Models}
    \label{fig:verdict_distribution}
\end{figure}

The reasoning strategies employed by LLMs also play a crucial role in their performance variability. Models in the o1 family leverage advanced CoT prompting, enabling logical, step-by-step reasoning that aligns with the structured problem-solving nature of ICPC challenges. This is evident in their ability to handle edge cases and adhere to strict computational constraints, as shown in Figure~\ref{fig:verdict_distribution}, where o1-mini achieves the highest percentage of ``Accepted" (AC) verdicts at 9.64\%.

In contrast, general-purpose models like GPT-4o and Llama-3.1-405B often lack CoT-specific training, resulting in a reliance on pattern recognition and memorized data. This limitation leads to higher error rates on unseen problems and significantly reduced adaptability, as reflected in their 0\% accuracy across all years \cite{b27}.

These findings reinforce the importance of specialized training methodologies and advanced reasoning strategies in achieving consistent and robust performance on competitive programming tasks. The combination of contamination-free datasets, tailored training, and effective reasoning strategies enables models like o1-mini and o1-preview to outperform general-purpose alternatives.
\\
\textbf{RQ4: Verdict Distribution Analysis:}
Verdict distributions highlight differences in error frequencies and success rates across models. As illustrated in Figure~\ref{fig:verdict_distribution}, o1-preview and o1-mini had the highest proportions of ``Accepted" (AC) verdicts, with 9.64\% for o1-mini and 9.04\% for o1-preview. In contrast, models like GPT-4o showed predominantly error states such as ``Compile Error" (CE) and ``Wrong Answer" (WA). Table~\ref{tab:verdict_distribution_counts} shows that o1-mini and o1-preview achieved 16 and 15 Accepted (AC) solutions, respectively, while GPT-4o, Mistral Large, and Llama-3.1-405B recorded 0 ACs. 

For error states, GPT-4o had the highest percentage of Compile Errors (CE) at 24.7\%, followed by Llama-3.1 at 19.88\% and Mistral Large at 18.07\%. ``Wrong Answer" (WA) verdicts were the most frequent error state across all models, with o1-mini recording 124 WA (74.70\%) and GPT-4o recording 55 WA (33.25\%).

These results emphasize the superiority of the o1 models, particularly their ability to achieve higher success rates and fewer Compile Errors compared to general-purpose models. However, the dominance of ``Wrong Answer" verdicts across all models highlights the challenges faced in generating accurate solutions for competitive programming tasks. Figure~\ref{fig:verdict_distribution} and Table~\ref{tab:verdict_distribution_counts} summarize these findings.

\begin{table}[htbp]
    \caption{Verdict Distribution Across Models (Counts)}
    \centering
    \begin{tabular}{@{}lcccccc@{}}
        \toprule
        \textbf{Model} & \textbf{AC} & \textbf{CE} & \textbf{RE} & \textbf{TLE} & \textbf{WA} & \textbf{MLE} \\
        \midrule
        o1-mini        & 16          & 10          & 0           & 13           & 124         & 1           \\
        o1-preview     & 15          & 19          & 2           & 10           & 120         & 0           \\
        GPT-4o         & 0           & 41          & 39          & 31           & 55          & 0           \\
        Mistral-Large  & 0           & 30          & 48          & 23           & 65          & 0           \\
        Llama-3.1      & 0           & 33          & 36          & 25           & 70          & 0           \\
        \bottomrule
    \end{tabular}
    \label{tab:verdict_distribution_counts}
\end{table}

These findings are consistent with \cite{b4}, which highlights the superior chain-of-thought reasoning capabilities of o1 models, enabling them to handle complex problem constraints more effectively.

\subsection{Key Insights}
The comprehensive evaluation highlights several key takeaways:
\begin{enumerate}
    \item \textbf{o1 Models Superior Performance}: o1-mini and o1-preview consistently outperform other LLMs in accuracy, verdict distribution, and resource efficiency, demonstrating their advanced reasoning and calibration.
    \item \textbf{Impact of Training Methodologies}: Models with specialized training for chain-of-thought reasoning exhibit greater robustness and adaptability to unseen problems.
    \item \textbf{Necessity for Contamination-Free Benchmarks}: The significant performance drop in general-purpose models on unseen data underscores the importance of uncontaminated benchmarks to accurately assess model generalization \cite{b2,b13}.
    \item \textbf{Resource Efficiency}: o1 models not only achieve higher accuracy but also demonstrate superior computational efficiency, making them more suitable for resource-constrained environments.
\end{enumerate}

These insights set the stage for future research in training methodologies and evaluation frameworks for LLMs in technical problem-solving.

\section{Threats to Validity}
In this section, we discuss four types of threats, similar to prior research \cite{pyreddy2025emoxpt}.
\subsection{Threats to Internal Validity}
Data contamination poses a major threat to our study’s internal validity. Some of the ICPC problems we used might already exist in the evaluated LLMs’ training data. We select
problems from years (2011-2024) that likely do not overlap with the training cutoffs of popular models, but ensuring no overlap is challenging. If any problem appears in the training data, they could artificially inflate performance scores, especially for models exposed to similar problem statements\cite{Xu2023Lemur}. Therefore, we evaluated the models on ICPC 2024's problems within 24 hours of the contest to ensure none of the models were trained on that data. 

\subsection{Threats to External Validity}
The study focuses on ICPC World Finals problems, which may limit the generalizability of the findings. While ICPC problems are diverse and complex, they represent only a subset of the programming challenges faced in broader software engineering and real-world application development. As a result, the performance of LLMs on ICPC problems may not directly translate to other programming tasks, potentially limiting the applicability of our conclusions to different domains \cite{b1,b2,b12}. To mitigate this bias, we use a diverse set of 166 ICPC World Finals problems, covering various categories and difficulty levels, to strengthen the robustness of the evaluation.

\subsection{Threats to Construct Validity}
We test the models in a zero-shot setting without additional fine-tuning or iterative interactions. This approach may not fully leverage the models’ capabilities, as fine-tuning or interactive prompting could improve performance on specific tasks. Additionally, differences in how each model interprets prompts or handles specific problem components may introduce variability not fully accounted for in our study. These constraints could limit the ability to generalize the findings to scenarios where models are fine-tuned or interactively guided during problem-solving \cite{b3,b4,b11}. To mitigate this bias, we document all experimental procedures and provide access to the scripts used for data preprocessing and solution evaluation to ensure reproducibility and transparency.
\subsection{Threats to Conclusion Validity}
Our evaluation relies exclusively on the Codeforces Gym platform for submitting solutions and determining verdicts. This creates a dependency on the platform’s specific execution environment and judging criteria. Variations in compiler versions, runtime environments, or hidden test cases used by the platform can affect the consistency and reliability of the verdicts. Additionally, platform-specific nuances in error reporting or resource measurement may influence the evaluation of resource utilization metrics \cite{b27,b13}. To minimize this bias, we ensure that all submissions are made under the same conditions within Codeforces Gym, maintaining consistency in verdict determination and resource measurement.
\ignore{
\subsection{Mitigation Strategies}
To address these threats, we employed several mitigation strategies:
\begin{itemize}
    \item \textbf{Data Selection}: Carefully selected problems from years less likely to overlap with the training data of major LLMs to minimize the risk of data contamination.
    \item \textbf{Diverse Problem Set}: Utilized a diverse set of 83 ICPC World Finals problems encompassing various categories and difficulty levels to enhance the robustness of the evaluation.
    \item \textbf{Consistent Evaluation Environment}: Ensured that all submissions were made under the same conditions within Codeforces Gym to maintain consistency in verdict determination and resource measurement.
    \item \textbf{Transparent Reporting}: Documented all experimental procedures and provided access to scripts used for data preprocessing and solution evaluation to facilitate reproducibility and transparency.
\end{itemize}

By acknowledging and addressing these threats, we aim to provide a balanced and comprehensive assessment of LLM performance on competitive programming benchmarks.
}
\section{Related Work}
Advancements in large language models (LLMs) have spurred significant interest in evaluating their capabilities across various domains. In competitive programming, several studies emphasize the importance of rigorous benchmarks and highlight challenges such as reasoning ability, data contamination, and evaluation methodologies.

\subsection{Competition-Level Problems as LLM Evaluators}
Programming challenges from platforms like Codeforces and the International Collegiate Programming Contest (ICPC) offer unique evaluation benchmarks for LLMs due to their complexity and diversity. These problems require a deep understanding of algorithms, mathematics, and reasoning, making them ideal for assessing LLM capabilities. Performance on unseen problems often drops significantly, indicating limitations in reasoning and generalization \cite{b1,b12}. These challenges underscore the need for reliable benchmarks and techniques to enhance reasoning in LLMs.

\subsection{Evaluation Methodologies}
Traditional methods for evaluating code generation have faced criticism due to issues like ``context leakage" and ``evolving-ignored" problems. Benchmarks such as those discussed in \cite{b2} better simulate real-world scenarios by considering the evolving nature of software development. These approaches reduce inflated performance metrics caused by unrealistic evaluation settings, providing a more accurate reflection of an LLM's problem-solving capabilities.

\subsection{Code Refinement and Interactivity}
Interactive code refinement and test-driven workflows have been shown to improve the quality of LLM-generated solutions. Iterative problem-solving techniques, combined with clear feedback, enhance LLM performance, particularly on complex programming challenges \cite{b3,b13}.

\subsection{Model-Specific Insights}
Studies focusing on specific models, such as the o1 family, emphasize their advanced chain-of-thought reasoning and robust handling of competitive programming problems \cite{b4}. These models outperform others in minimizing hallucinations and achieving consistent performance across diverse tasks, further validating their efficacy for high-stakes evaluations. Similarly, Mistral 7B demonstrates optimized performance for efficiency, highlighting the importance of model design for resource-constrained environments \cite{b11}.

\subsection{Challenges in Code Evaluation}
Issues like data contamination, overfitting, and reliance on pretraining data limit the generalization of LLMs. Research on the calibration and correctness of LLM-generated code highlights the importance of confidence scores and error analysis in ensuring reliable outputs, particularly in scenarios requiring high precision \cite{b6}.

\subsection{Performance Comparisons}
Comparative studies of models like GPT-4o, Llama-3.1, and o1 systems reveal stark differences in accuracy and resource efficiency \cite{b27,b10}. These comparisons underscore the importance of both training data and architectural design in achieving superior results on competitive programming problems.

By synthesizing these findings, our work contributes to the growing body of research by evaluating multiple LLMs on ICPC-style problems. This expands on previous studies by combining problem-solving performance with detailed error and resource utilization analysis.

\section{Conclusion}
This study underscores the efficacy of competition-level programming problems, specifically those from the International Collegiate Programming Contest (ICPC), as robust benchmarks for evaluating large language models (LLMs). Utilizing a curated dataset of 166 ICPC World Finals problems from 2011 to 2024, we systematically assessed state-of-the-art LLMs, including OpenAI's o1 family, GPT-4o, Mistral Large, and Llama-3.1-405B. Our findings reveal that the o1 models, particularly o1-mini and o1-preview, significantly outperform others in terms of accuracy, robustness, and computational efficiency, owing to their advanced chain-of-thought (CoT) reasoning and diverse training methodologies. In contrast, models such as GPT-4o, Mistral Large, and Llama-3.1-405B demonstrate limited generalization abilities and higher error rates on unseen 2024 problems, highlighting their reliance on pretraining data and the need for improved generalization skills \cite{b1,b6}. Additionally, the analysis of verdict distributions indicates that o1 models consistently achieve higher proportions of ``Accepted" (AC) verdicts while minimizing errors like ``Compile Error" (CE) and ``Time Limit Exceeded" (TLE), in contrast to other models which show higher frequencies of these errors. These insights emphasize the importance of developing contamination-free benchmarks and enhancing reasoning capabilities through advanced training methodologies like CoT and iterative refinement \cite{b3,b4}. Overall, \Name{} contributes to a deeper understanding of LLM performance in real-world problem-solving scenarios and paves the way for further advancements in LLM design and evaluation, ensuring that they can meet the complex demands of technical problem-solving tasks.

\bibliographystyle{IEEEtran}
\bibliography{main}

\end{document}